\documentclass{article}

    \PassOptionsToPackage{numbers, compress}{natbib}

\usepackage[preprint]{neurips_2025}

\usepackage[utf8]{inputenc} 
\usepackage[T1]{fontenc}    
\usepackage{hyperref}       
\usepackage{url}            
\usepackage{booktabs}       
\usepackage{amsfonts}       
\usepackage{nicefrac}       
\usepackage{microtype}      
\usepackage{xcolor}         

\usepackage{amsmath} 

\usepackage{graphicx}

\usepackage{algorithm}

\usepackage{algorithmic}

\usepackage{comment}

\usepackage{tcolorbox}
\usepackage{xcolor}
\usepackage{tcolorbox}
\usepackage{listings}
\usepackage{fancyvrb}
\tcbuselibrary{listings}

\newcommand{\bbbeta}{\boldsymbol{\beta}}
\newcommand{\bbmu}{\boldsymbol{\mu}}

\newcommand{\bbSig}{\boldsymbol{\Sigma}}
\newcommand{\bbPhi}{\boldsymbol{\Phi}} 
\newcommand{\bbphi}{\boldsymbol{\phi}} 
\newcommand{\change}[1]{{\color{red}#1}}

\title{Fine-tuning LLMs with variational Bayesian last layer \\for high-dimensional Bayesian optimization}

\author{
  Haotian Xiang \\
  School of ECE \\
  University of Georgia \\
  Athens, GA \\
  \texttt{Haotian.Xiang@uga.edu} \\
  \And
  Jinwen Xu \\
  School of ECE \\
  University of Georgia \\
  Athens, GA \\
  \texttt{Jinwen.Xu@uga.edu} \\
  \AND
  Qin Lu\thanks{Corresponding author: \texttt{Qin.Lu@uga.edu}}  \\
  School of ECE \\
  University of Georgia \\
  Athens, GA \\
  \texttt{Qin.Lu@uga.edu} \\
}

\begin{document}

\maketitle

\begin{abstract}
A plethora of applications entail solving {\it black-box} optimization problems with high evaluation costs, including drug discovery, material design, as well as hyperparameter tuning. Toward finding the global optimum of such black-box optimization problems with sample efficiency, Bayesian optimization (BO) is a theoretically elegant framework that relies on a probabilistic surrogate model so as to iteratively select the query point with well-balanced exploration-exploitation tradeoffs. The Gaussian process (GP), as the de-facto choice for surrogate modeling, has achieved compelling performances for {\it vanilla} BO with low-dimensional continuous variables. However, GPs fall short in coping with high-dimensional counterparts with {\it irregular} variables (e.g., categorical, ordinal, etc.). To alleviate this, neural network-based surrogates have been explored.
Inspired by the powerful capabilities of LLMs, we adopt the LLM as the surrogate to model the mapping from the high-dimensional input variables to the objective function. To adapt to the current problem, we leverage the low-rank adaptation (LoRA) to fine-tune the LLM parameters together with the posterior of a linear regression head via the variational Bayesian last layer (VBLL) framework. The resulting LoRA-VBLL is not only computationally light compared to existing alternatives, but also admits recursive updates. To automate the critical selection of the LoRA rank as well as other hyperparameters, a weighted ensemble (ENS) of LoRA-VBLL surrogates has been devised, which further accommodates continual update of the per-model weight and individual LoRA-VBLL parameters via recursive Bayes. Extensive experimental results demonstrate the compelling performance of the proposed (ENS-)LoRA-VBLL approaches on various high-dimensional benchmarks and the real-world molecular optimization tasks.
\end{abstract}

\section{Introduction}
A number of applications in science and engineering amount to optimizing an `expensive-to-evaluate' black-box function, including hyperparameter tuning \cite{snoek2012practical}, drug discovery \cite{korovina2020chembo}, and policy optimization in robotics\cite{cully2015robots}. As in hyperparameter tuning, lack of analytic expressions for the objective function and overwhelming evaluation cost discourage grid search and adoption of gradient-based solvers. To find the global optimum with a limited evaluation budget, Bayesian optimization (BO) offers a principled framework by leveraging a statistical model to guide the acquisition of query points on-the-fly, conforming to the well-known exploration-exploitation tradeoff~\cite{garnett2023bayesian}.

Thanks to its well-calibrated uncertainty estimates and sample efficiency, the Gaussian process (GP) is the most widely adopted surrogate model for BO. Although GP-based BO yields competitive performances when the decision variables are low-dimensional and continuous and the learning function is stationary, it falls short in dealing with high-dimensional and nonstationary spaces with irregular variables (e.g., categorical or ordinal). In molecular optimization, for example, the search space is inherently discrete, high-dimensional, and contains complex structural dependencies that are difficult to capture with standard GP kernels. To alleviate this challenge, Bayesian neural network (BNN) based surrogates have gained increasing popularity; see, e.g., ~\cite{snoek2015scalable,springenberg2016bayesian, li2024study}. While BNNs are adept at accommodating nonstationarity and inherently scale with data, efficient approximate posterior inference with robustness to under-fitting is still a challenge. Combining the merits of BNNs and GPs, Bayesian last layer models (BLL) rely on {\it deterministic} NN-based feature mapping and linear random regression head. Upon variational (V) training, the resulting VBLL approach jointly optimizes NN weights and the posterior model, yielding compelling performances in various BO benchmark tasks~\cite{brunzema2024bayesian}.

Recently, the advent of ChatGPT has sparked a lot of research endeavors of unleashing the power of pre-trained large language models (LLMs) for various application domains; see, e.g., ~\cite{thirunavukarasu2023large, bommasani2021opportunities}. This prompts a natural question: can we leverage LLMs as surrogate models for BO to overcome the limitations of traditional approaches? The high-dimensional embedding spaces of LLMs and their inherent ability to handle irregular variables make them promising candidates for addressing the challenges of high-dimensional BO. However, to fully realize their potential in BO, we need efficient and reliable methods to quantify uncertainty in LLM predictions, which is essential for the exploration-exploitation trade-off in acquisition functions.

Several recent attempts have been made to leverage LLMs to enhance the performance of BO via either in-context learning~\cite{liu2024large,ramos2023,yin2025adollm}, or surrogate modeling~\cite{kristiadi2024sober}.  However, existing approaches face limitations. The Laplace approximation (LA), while computationally efficient, provides only a post-hoc uncertainty estimation that may lead to suboptimal results~\cite{ritter2018scalable}. The more recent Bayesian Low-rank adaptation by Backpropagation (BLoB) approach performs variational inference over the Low-Rank Adaptation (LoRA) parameters~\cite{hu2022lora}, but requires Monte Carlo sampling to compute the likelihood term in the Evidence Lower Bound (ELBO), resulting in excessive memory consumption and computational overhead~\cite{wang2024blob}.

\noindent {\bf Contributions.} The contributions of the current work are summarized as follows.

\begin{itemize}
  \item[c1)] We propos\textit{}e a novel approach that adapts VBLL to LLM backbones for high-dimensional BO. Our method leverages the computational efficiency of LoRA for parameter-efficient fine-tuning while relying on the VBLL framework for scalable Bayesian uncertainty estimation. 
\item[c2)] The proposed LoRA-VBLL is amenable to recursive Bayes updates with scalability. When the optimization variables are selected from a pool of candidates, one can further leverage feature caching, significantly reducing computational demands compared to other methods. 
  \item[c3)] To automate the selection of the LoRA rank, we employ an ensemble (ENS) of LoRA-VBLL models, each of which is associated with a unique rank and is assessed via data-adaptive weight.  The resulting ENS-LoRA-VBLL approach also accommodates recursive updates and feature caching.
    
    
  \item[c4)] Comprehensive experiments across various high-dimensional BO tasks demonstrate that the proposed (ENS)-LoRA-VBLL consistently outperform existing methods in both optimization performance and computational efficiency.
\end{itemize}

\section{Preliminaries and related works}
A plethora of applications can be mathematically abstracted as the following  optimization problem
\begin{align}
     {\bf x}_* = \underset{{\bf x}\in{\cal X}}{\arg\max} \ f({\bf x}) \label{eq:f}
\end{align}
where ${\bf x}$ is the sought $n_x$-dimensional optimization variables with feasible set ${\cal X}$, and $f$ is the {\it black-box} objective which has no analytic expression and is expensive to evaluate. Towards solving~\eqref{eq:f} in a sample-efficient manner, BO offers a theoretically elegant solution by {\it actively} selecting informative query pairs for a given evaluation budget~\cite{garnett2023bayesian}.
The most fundamental step in BO is to infer a {\it stochastic} belief for the function $f({\bf x})$  using the evaluated input-output pairs  $\mathcal{D}_t:= \{({\bf x}_\tau, y_\tau)\}_{\tau=1}^t$, based on which $\mathbf{x}_{t+1}$ is selected {\it actively}. Specifically, this procedure is implemented iteratively via two steps, namely, \textbf{s1)}  obtain  $p(f({\bf x})|\mathcal{D}_t)$ based on the surrogate model; and \textbf{s2)} select ${\bf x}_{t+1}\! =\! {\arg\max}_{{\bf x}\in {\cal X} }  \ \alpha ({\bf x}|\mathcal{D}_t)$,  where the acquisition function (AF) $\alpha$, usually available in closed form, is designed based on $p(\! f(\mathbf{x})|\mathcal{D}_t)$ to strike a balance between {\it exploration} and {\it exploitation}. 

For the {\it vanilla} BO setting where the optimization variables are low-dimensional and continuous, the GP, thanks to its well-calibrated uncertainty quantifiability and interpretability, is the default choice for surrogate modeling. Based on different design rules, a suite of GP-based AFs has been derived~\cite{garnett2023bayesian}. 
In spite of the compelling performance, GP-based BO falls short in coping with high-dimensional variables since it is challenging to properly fit a GP surrogate without proper regularization ~\cite{hvarfner2024vanilla, xu2024standard} and optimizing a high-dimensional AF is also nontrivial. This issue is further exacerbated when there are categorical, ordinal, or mixed variables in the input space due to the challenge of capturing the meaningful pairwise distance in the kernel function.

\subsection{Related Works}
To put the current contribution in context, the following existing works will be outlined.

\noindent {\bf High-dimensional (HD) BO over categorical or mixed-variable spaces.} 
Previous attempts towards HDBO typically enforce structural assumptions of the surrogate model, either in the function or input space. For the former, the common practise is to seek a sum of functions, whose input variables belong to a low-dimensional subset of the HD input space\cite{kandasamy2015high}. Alternatively, one can resort to the  {\it latent} BO approaches, where, instead of optimizing the HD variables, the low-dimensional projections are selected via standard BO methods. There are several developments to obtain the mapping, including random embedding~\cite{wang2016bayesian}, non-linear projection based on the variational auto-encoders~\cite{gomez2018automatic}, or structured projection~\cite{jaquier2020high}.
Meanwhile, efforts have been made for BO over categorical and mixed variables~\cite{dreczkowski2023framework}. Early methods developed specialized kernel functions, with COMBO~\cite{oh2019combinatorial} introducing diffusion kernels using graph representations of the search space. BOSS~\cite{moss2020boss} and BOiLS~\cite{grosnit2022boils} leveraged string subsequence kernels, effective for problems where categorical order matters. For mixed spaces, CoCaBO~\cite{ru2020bayesian} combined multi-armed bandits with GPs using additive kernels. Casmopolitan~\cite{wan2021think} advanced this with trust regions and hybrid acquisition optimization. Alternative approaches include BOCS~\cite{baptista2018bayesian} using Bayesian linear regression with horseshoe priors for better computational efficiency, and BODi~\cite{deshwal2023bayesian} employing dictionary-based embeddings for HD problems. NN-based surrogates have also been explored~\cite{lee2017deep}, despite uncertainty quantification challenges. 
Please refer to~\cite{gonzalez2024survey, dreczkowski2023framework} for a comprehensive review.

\noindent {\bf BNNs and NN-based BO.}  Bayesian approaches are predominant for uncertainty reasoning over deep NNs. A variety of approximate inference techniques have been developed, ranging from high-quality  Hamiltonian Monte Carlo~\cite{neal2012bayesian}, to stochastic gradient Markov Chain Monte Carlo~\cite{chen2014stochastic}, to Monte Carlo dropout~\cite{gal2016dropout}, as well as heuristics such as deep ensembles~\cite{lakshminarayanan2017simple}. Combining the merits of BNNs and GPs, hybrid approaches have also been devised, including deep kernel learning (DKL)~\cite{wilson2016deep} that leverages deterministic NN-based feature extractor as the input to the GP model, and infinite-width BNN~\cite{lee2018deep}, which boils down to a GP with the kernel constructed by layer-wise composition as in NNs. A study of these BNN-based BO has been conducted in~\cite{li2024study}. More recently, another hybrid approach is the variational Bayesian last layers, which, similar to DKL, stacks an NN-based feature vector with a linear random regression head. Unlike DKL that relies on the marginal likelihood for model training, VBLL operates by variational training, that jointly seeks the model parameters and the model posterior pdf. This lighweight Bayesian method is shown to outperform other NN-based baselines for the BO task in~\cite{brunzema2024bayesian}. In addition to BNN, another line of work uses deterministic training to obtain NN weights and then infer uncertainty values in a post-hoc manner around the estimated parameters; see, e.g., ~\cite{zhou2020neural,zhang2021neural}.

\noindent {\bf LLMs for BO and Bayesian fine-tuning of LLMs.} Inspired by the striking performances of LLMs, recent works have explored integrating LLMs to enhance BO. The first attempt is the direct in-context learning (ICL), which, without updating LLM parameters, makes inferences based on examples within a prompt~\cite{liu2024large,ramos2023,yin2025adollm}. Thanks to LLMs' few-shot learning capabilities, such as ICL, can empirically enhance the performance of BO-based hyperparameter tuning via zero-shot warmstarting, candidate points sampling, as well as surrogate model mapping~\cite{liu2024large}.
In spite of this, they often struggle to maintain consistent performance across diverse tasks and require careful prompt engineering to elicit proper uncertainty estimates. To alleviate these issues, LLM-based surrogate modeling for BO has been explored in~\cite{kristiadi2024sober}, where it is shown that parameter-efficient fine-tuning of LLMs with uncertainty estimates is critical. The most straightforward route to this is via Laplace approximation (LA), which is a computationally efficient way to obtain uncertainty estimates after fine-tuning LLM parameters~\cite{yang2023bayesian}. Not specifically for the BO context, an alternative Bayesian low-rank adaptation by Backpropagation (BLoB) has been devised in~\cite{wang2024blob}, and is shown to outperform the LA-based counterpart in terms of generalization and uncertainty estimation. 
However, this is at the expense of increased computational complexity. Here, we build on the VBLL framework to devise a computationally lightweight Bayesian fine-tuning framework, which will not only benefit the BO task here, but also other prediction-oriented tasks in general.


\section{LLM-enhanced surrogate model for HDBO}

Towards addressing the challenging HDBO task with irregular variables~\eqref{eq:f}, we will adapt pre-trained LLMs as the surrogate model thanks to their powerful feature representations and zero-/few-shot learning capabilities. To proceed, we will convert the non-language BO task to a language task through LIFT~\cite{dinh2022lift}. Please refer to Appendix~A for the details.

\subsection{Fine-tuning LLMs via variational Bayes last layer}
Adapting model parameters of LLMs efficiently while maintaining performance is a crucial challenge. Among various attempts to fine-tune LLMs in a parameter-efficient fashion~\cite{mangrulkar2022peft}, we will leverage the Low-Rank Adaptation (LoRA), which, rather than updating all model weights directly, decomposes weight updates into low-rank representations~\cite{hu2022lora}. Specifically, given pre-trained LLM weights ${\bf W}_0\in \mathbb{R}^{m\times n}$, LoRA yields the fine-tuned weights given by ${\bf W} := {\bf W}_0 + {\bf A}^\top {\bf B}$, where $\mathbf{B} \in \mathbb{R}^{r \times n}$ and $\mathbf{A} \in \mathbb{R}^{r \times m}$ are low-rank matrices with rank $r \ll \min(m,n)$. Let 
$\bbphi_{{\bf W}_0+{\bf A}^\top {\bf B}}\in \mathbb{R}^d$ be the $d$-dimensional feature mapping with fine-tuned weights. Here, we will rely on the following Bayesian last-layer (BLL) surrogate model
\begin{equation}
    y_t =  \bbphi^{\top}_{{\bf W}_0+{\bf A}^\top {\bf B}} ({\bf x}_t)\bbbeta + \epsilon_t, \epsilon_t\sim{\cal N}(0, \sigma_{\epsilon}^2)\label{eq:model}
\end{equation}
where $\bbbeta$ is the random linear regression head with Gaussian prior $p({\bbbeta}) = {\cal N}({\bbbeta}; {\bf 0}, \sigma_\beta^2 {\bf I}_d)$ with variance $\sigma_\beta^2$. Alternatively, one can express~\eqref{eq:model} via the likelihood $p(y_t|\bbbeta, {\bf x}_t; {\bf A}, {\bf B},\sigma_\epsilon^2) = {\cal N}(y_t; {\bbphi}^\top_{{\bf W}_0+{\bf A}^\top {\bf B}} ({\bf x}_t) {\bbbeta}, \sigma_\epsilon^2)$.

Given data collected in ${\cal D}_t$, such a BLL model has a closed-form expression of the marginal likelihood,
based on which one can leverage gradient descent to obtain the LoRA parameters $\{{\bf A}, {\bf B}\}$, along with other hyperparameters. However, this is not only computationally prohibitive, but also numerically unstable~\cite{harrison2018meta}. 
Towards addressing these issues, we will leverage variational inference to jointly learn the LoRA parameters $\{{\bf A}, {\bf B}\}$ and the posterior of $\bbbeta$. This is also termed variational (V) BLL training~\cite{brunzema2024bayesian}. Hereafter, we will term the proposed approach ``LoRA-VBLL". Let the posterior of ${\bbbeta}$ be denoted as $q_t(\bbbeta) = \mathcal{N}(\bbbeta; \boldsymbol{\mu}_t, \boldsymbol{\Sigma}_t)$. Then, $\boldsymbol{\mu}_t$ and $\boldsymbol{\Sigma}_t$ are sought together with $\{{\bf A}, {\bf B}\}$ and $\sigma_\epsilon^2$ by maximizing the evidence lower bound (ELBO) ($\boldsymbol{\Theta}:=\{\boldsymbol{\mu}_t,\boldsymbol{\Sigma}_t,{\bf A},{\bf B}, \sigma_\epsilon^2 \}$) using mini-batch stochastic gradient descent (SGD)
\begin{align}
   \hat{\boldsymbol{\Theta}} = \underset{\boldsymbol{\Theta}}{\arg\max} \ {\cal L}^{\rm ELBO}_t(\boldsymbol{\Theta})\label{eq:max_elbo}
\end{align}
where ${\cal L}_t^{\rm ELBO}(\boldsymbol{\Theta})$, as a lower bound to the log-margninal likelihood $\log p({\cal D}_t; \boldsymbol{\Theta})$, is given explictly as
\begin{align}
{\cal L}_t^{\rm ELBO}(\boldsymbol{\Theta})
&= \mathbb{E}_{q_t({\bbbeta})} \sum_{t}[\log (p(y_t|{\bbbeta}, {\bf x}_t; {\bf A}, {\bf B},\sigma_\epsilon^2)] - {\rm KL}(q_t({\bbbeta})\| p({\bbbeta}))  \nonumber\\
& =\  -\frac{1}{2}\log\left(2\pi\sigma_\epsilon^2\right) -\frac{1}{2\sigma_\epsilon^2}\sum_{t}\left[
    \left(y_t - \boldsymbol{\phi}^{\top}_{\mathbf{W}_0 + \mathbf{A}^{\top}\mathbf{B}}\left(\mathbf{x}_t\right)\boldsymbol{\mu}_t\right)^2
  \right] \nonumber\\
  &\ \quad -\frac{1}{2\sigma_\epsilon^2}\sum_{t}
    \boldsymbol{\phi}^{\top}_{\mathbf{W}_0 + \mathbf{A}^{\top}\mathbf{B}}\left(\mathbf{x}_t\right)\boldsymbol{\Sigma}_t
    \boldsymbol{\phi}_{\mathbf{W}_0 + \mathbf{A}^{\top}\mathbf{B}}\left(\mathbf{x}_t\right)
\nonumber \\
&\ \quad -\frac{1}{2\sigma_{\beta}^2}\left(\operatorname{Tr}\left(\boldsymbol{\Sigma}_t\right) + \boldsymbol{\mu}_t^{\top}\boldsymbol{\mu}_t\right) +\frac{1}{2}\left[k + \log\frac{\left|\boldsymbol{\Sigma}_t\right|}{\sigma_{\beta}^{2k}}\right]\;. \label{eq:ELBO}
\end{align}

Given the posterior $q_t(\bbbeta)$ after the VBLL training in ~\eqref{eq:max_elbo}, one can obtain the  function predictive posterior as
\begin{align}
    p(f({\bf x})|{\cal D}_t; \hat{\boldsymbol{\Theta}}) = {\cal N}(f({\bf x}); {\bbphi}^\top_{\small {\bf W}_0+ \hat{\bf A}^\top \hat{\bf B}} ({\bf x}) \hat{\bbmu}_t, {\bbphi}^\top_{{\bf W}_0+\hat{\bf A}^\top \hat{\bf B}} ({\bf x})\hat{\bbSig}_t {\bbphi}_{{\bf W}_0+ \hat{\bf A}^\top \hat{\bf B}} ({\bf x}) ) \label{eq:f_pre}
\end{align}
Based on which, we are ready to design the AF. Here, we will adopt the Thompson sampling (TS) in the experiments thanks to the parametric model and the convenience in incorporating batch evaluation
\begin{align}
   \alpha ({\bf x}|\mathcal{D}_t) = \bbphi^\top_{{\bf W}+ \hat{\bf A}^\top \hat{\bf B}} ({\bf x}) \tilde{\bbbeta}_t,\quad  \tilde{\bbbeta}_t \sim q_t(\bbbeta) \label{eq:AF}
\end{align}
Towards optimizing the HD AF~\eqref{eq:AF}, we will adopt the trust region-based local search approach~\cite{wan2021think}. This approach utilizes a hill-climbing-based local search method that restricts the search to promising regions surrounding the current best solution, thus effectively balancing local exploration and exploitation. For categorical variables, the trust region boundaries are defined by Hamming distance, while continuous variables are constrained within hyper-rectangular regions that adaptively expand or contract based on the success of the optimization efforts. 

\subsection{Recursive incremental update between fine-tuning}
After acquiring the label $y_{t+1}$ of ${\bf x}_{t+1}$, we are ready to update the surrogate model. Fine-tuning in every iteration may not necessarily yield improved performance and has much increased complexity. Thus, we choose to update LoRA parameters in every a few iterations. Between any two neighboring updates, we will implement an incremental update for the posterior of $\bbbeta$ via recursive Bayes. As both the posterior and the likelihood are Gaussian, this will yield a nice closed-form update. Specifically, given a newly acquired pair $({\bf x}_{t+1}, y_{t+1})$, the posterior pdf of $\bbbeta$ can be propagated as
\begin{align}
	q_{t+1}(\bbbeta)  & =  	\frac{q_t(\bbbeta) {p}(y_{t+1}|\bbbeta,\mathbf{x}_{t+1};\hat{\bf A}, \hat{\bf B},\hat{\sigma}_\epsilon^2)}
	{{p}(y_{t+1}|\mathbf{x}_{t+1}, \mathcal{D}_{t})} = \mathcal{N}(\bbbeta; \hat{\bbmu}_{t+1}, \hat{\bbSig}_{t+1})\label{eq:beta_up}
\end{align}
where the updated mean $ \hat{\bbmu}_{t+1}$ and covariance matrix $\bbSig_{t+1}$ are ($\hat{\bbphi}:=\bbphi_{{\bf W}+ \hat{\bf A}^\top \hat{\bf B}}$)
\begin{subequations} \label{eq:posterior_update_1}
	\begin{align}
		\hspace*{-0.25cm}\hat{\bbmu}_{t+1} &= \hat{\bbmu}_{t} +  \sigma_{t+1|t}^{-2}\hat{\bbSig}_{t} \hat{\bbphi}(\mathbf{x}_{t+1})(y_{t+1} - \hat{y}_{t+1|t}) \nonumber\\
		\hspace*{-0.25cm}	\hat{\bbSig}_{t+1} &= \hat{\bbSig}_{t} - \sigma_{t+1|t}^{-2}\hat{\bbSig}_{t} \hat{\bbphi}(\mathbf{x}_{t+1}) \hat{\bbphi}^{\top}(\mathbf{x}_{t+1})\hat{\bbSig}_{t}. \nonumber
	\end{align}
\end{subequations}
with $\hat{y}_{t+1|t} =  \hat{\bbphi}^{\top} \!(\mathbf{x}_{t+1})\hat{\bbmu}_t$ and $\sigma_{t+1|t}^2 = \hat{\bbphi}^{\top} (\mathbf{x}_{t+1}) \hat{\bbSig}_{t} \hat{\bbphi}(\mathbf{x}_{t+1})+\hat{\sigma}_\epsilon^2$.

\noindent{\bf Remark 1.} When to fine-tune the LoRA parameters is a critical choice. Here, we will adopt the strategy with event-based trigger, which chooses to fine-tune when the predictive likelihood $p(y_{t+1}|{\bf x}_{t+1}, {\cal D}_t)$ is less than a pre-defined threshold~\cite{brunzema2024bayesian}. Please also refer to the ablation study in Sec.~5.3.2.

\noindent{\bf Remark 2.} As the forward passes of LLMs are computationally expensive, we can cache the LLM features between two fine-tuning steps if ${\bf x}$ is chosen from a pool of candidates, as in molecular optimization~\cite{kristiadi2024sober}. This is significantly more scalable than other Bayesian fine-tuning methods, including LA~\cite{yang2023bayesian} and BLoB~\cite{wang2024blob}.


\section{Ensembling LoRA-VBLL with recursive updates}
In the LoRA-VBLL approach, choosing the rank $r$ of the LoRA parameters is critical. To automate the selection of $r$, we here adopt an ensemble of $I$ LoRA-VBLL models, each associated with a unique rank $r_j$ ($j\in {\cal J}:=\{1,\ldots, J\}$) selected from a prescribed dictionary ${\cal R}$. Let ${\bf A}_j\in \mathbb{R}_{r_j\times M}$ and ${\bf B}_j\in \mathbb{R}_{r_j\times N}$  be the LoRA parameters associated with rank $r_j$, $\sigma_{\beta_j}^2$ be the prior covariance for the regression head $j$. The generative model for LoRA \textcolor{blue}{$j$} can be characterized through 
\begin{align}
	p(y_t|\bbbeta_j, \mathbf{x}_t) =
	\mathcal{N}(y_t; \bbphi_{{\bf W}_0 + {\bf A}_j^\top {\bf B}_j}^{\top} (\mathbf{x}_t)\bbbeta_j, \sigma_{\epsilon_j}^2) \quad \bbbeta_j\sim {\cal N}({\bf 0}, \sigma_{\beta_j}^2)\;.\label{eq:LF}
\end{align}
Similar to a single LoRA-VBLL model in the previous section, for each LoRA-BLL model, we will again rely on the ELBO objective to jointly optimize the model parameters $\{ {\bf A}_j, {\bf B}_j, \sigma_{\epsilon_j}^2 \}$ with the posterior $q_t(\bbbeta_j) = \mathcal{N}(\bbbeta; \boldsymbol{\mu}^j_t, \boldsymbol{\Sigma}_t^j)$. With $\boldsymbol{\Theta}
_j:=\{\boldsymbol{\mu}^j_t,\boldsymbol{\Sigma}^j_t,{\bf A}_j,{\bf B}_j, \sigma_{\epsilon_j}^2 \}$, the variational fine-tuning is given by
\begin{align}
   \hat{\boldsymbol{\Theta}}_j = \underset{\boldsymbol{\Theta}}{\arg\max} \ {\cal L}^{\rm ELBO}_t(\boldsymbol{\Theta}_j)\label{eq:max_elbo}\;.
\end{align}
Similar to~\eqref{eq:f_pre}, the function posterior predictive pdf per LoRA-VBLL is  then given by
\begin{align}
       p(f({\bf x})|{\cal D}_t, j; \hat{\boldsymbol{\Theta}}_j) = {\cal N}(f({\bf x}); \hat{\bbphi}^\top_j ({\bf x}) \hat{\bbmu}_t^j, \hat{\bbphi}_{j}^\top ({\bf x})\hat{\bbSig}^j_t \hat{\bbphi}_{j} ({\bf x}) )\label{eq:ens_fun}
\end{align}
where  $\hat{\bbphi}_{j}:=\bbphi_{{\bf W}_0 + \hat{\bf A}_j^\top \hat{\bf B}_j}$ for notational brevity.

One can then leverage the sum-product rule to yield the marginal function predictive posterior given by the following Gaussian mixture (GM)~\cite{polyzos2023bayesian,lu2023surrogate,lu2022incremental}
\begin{align}
	{p}(f(\mathbf{x})|\mathcal{D}_{t}; \{\hat{\boldsymbol{\Theta}}_j\}) \! &=\! \sum_{j = 1}^J\! {\rm Pr}(j|\mathcal{D}_{t}; \hat{\boldsymbol{\Theta}}_j) {p}(f(\mathbf{x})|\mathcal{D}_{t},j; \hat{\boldsymbol{\Theta}}_j) \label{eq:EGP_post}
\end{align}
where the posteior weight $w_t^j:={\rm Pr}(j|\mathcal{D}_{t}; \hat{\boldsymbol{\Theta}}_j)$ can be further factorized via the Bayes' rule as
\begin{align}
    w_t^j \propto w_0^j p({\cal D}_t|j; \hat{\boldsymbol{\Theta}}_j) \;.\label{eq: weight}
\end{align}
Here, the prior weight $w_0^j$ can be set to $1/J$ without any prior informatin, and $p({\cal D}_t|j; \hat{\boldsymbol{\Theta}}_j)$ is the marginal likelihood that is equivalent to the optimal ELBO value conditioned on $\hat{\boldsymbol{\Theta}}_j$ since the variational and true posterior of $\bbbeta_j$ are both Gaussian~\cite{brunzema2024bayesian}. That is,
\begin{align}
   p({\cal D}_t|j; \hat{\boldsymbol{\Theta}}_j) =  \exp({\cal L}^{\rm ELBO}_t(\hat{\boldsymbol{\Theta}}_j))\;. \label{eq: weight_ELBO}
\end{align}

Given~\eqref{eq:ens_fun}, we will again leverage the TS, which entails sampling from ~\eqref{eq:EGP_post} to acquire the next evaluation point, in accordance with~\cite{lu2023surrogate}. Given $\mathcal{D}_t$, acquisition of $\mathbf{x}_{t+1}$ is obtained as the maximizer of the random function sample $\tilde{f}_t (\mathbf{x})$, whose detailed implementation is given by
\begin{align}
	 \mathbf{x}_{t+1} &= \underset{\mathbf{x}\in\mathcal{X}}{\arg\max} \   \hat{\bbphi}_{j_t}^{ \top} (\mathbf{x}) \tilde{\bbbeta}_t   \quad {\rm where}\ {j}_t \sim {\cal CAT}(\mathcal{J},\mathbf{w}_t),   \ \tilde{\bbbeta}_t \sim p(\bbbeta_{j_t }|\mathcal{D}_t) \label{eq:ensemble_AF}
\end{align}
where ${\cal CAT}(\mathcal{J},\mathbf{w}_t)$ refers to a categorical distribution that samples values from $\mathcal{J}$ based on the probability vector $\mathbf{w}_t$. As with~\eqref{eq:AF}, ~\eqref{eq:ensemble_AF} can be readily solved via the trust region-based search algorithm in~\cite{wan2021think}.


\subsection{Recursive update}
As with the single LoRA-VBLL model, it is not efficient to update LoRA parameters of the ENS-LoRA-VBLL in every iteration. Thus, we will fine-tune LoRA parameters adaptively. Between any two fine-tuning steps, one can leverage recursive Bayes to update $\{w_t^j, \hat{\bbmu}_t^j, \hat{\bbSig}_t^j, j\in\mathcal{J}\}$ from slot to slot with scalability. For notational brevity, we will drop the dependence on $\{\hat{\boldsymbol{\Theta}}_j\}$ henceforth.

Specifically, upon acquiring the evaluation output $y_{t+1}$ for the selected input $\mathbf{x}_{t+1}$, the updated weight $w_{t+1}^j := {\rm Pr}(j|\mathcal{D}_{t},\mathbf{x}_{t+1}, y_{t+1})$ can be obtained via Bayes' rule as
\begin{align}
	w_{t+1}^j &
	\!	=\! \frac{{\rm Pr}(j|\mathcal{D}_{t},\mathbf{x}_{t+1}\!) {p}(y_{t+1}|\mathbf{x}_{t+1},  j,\mathcal{D}_{t})}{{p}(y_{t+1}|\mathbf{x}_{t+1}, \mathcal{D}_{t})} =\!   \frac{w_t^j \mathcal{N}\!\left(y_{t+1};  \hat{y}_{t+1|t}^{j}, (\sigma_{t+1|t}^{j})^2 \right)}{\sum_{j' = 1}^J w_t^{j} \mathcal{N}\!\left(y_{t+1};  \hat{y}_{t+1|t}^{j'}, \!(\sigma_{t+1|t}^{j'})^2 \!\right)}\label{eq:w_update} 	\end{align}
where the sum-product rule allows one to obtain the per-Gaussian predictive likelihood as 
\begin{align}
    {p}(y_{t+1}|j,\mathcal{D}_{t}, \mathbf{x}_{t+1}) = \int_{\bbbeta_j} p(y_{t+1}|\bbbeta_j, {\bf x}_{t+1})q_t(\bbbeta_j) d\bbbeta_j = \mathcal{N}(y_{t+1};\hat{y}_{t+1|t}^{j},(\sigma_{t+1|t}^{j})^2)
\end{align}
with $\hat{y}_{t+1|t}^{j} =  \hat{\bbphi}^{\top}_{j} (\mathbf{x}_{t+1})\hat{\bbmu}_t^{j}$ and $(\sigma_{t+1|t}^{j})^2 = \hat{\bbphi}^{\top}_{j}  (\mathbf{x}_{t+1}) \hat{\bbSig}^j_{t} \hat{\bbphi}_j (\mathbf{x}_{t+1})+\hat{\sigma}_{\epsilon_j}^2$.

Further, the posterior pdf of $\bbbeta_j$ can be propagated in the recursive Bayes fashion as
\begin{align}
	q_{t+1}(\bbbeta_j)  & =  	\frac{q_t(\bbbeta_j) {p}(y_{t+1}|\bbbeta_j,\mathbf{x}_{t+1}; \hat{\bf A}_j, \hat{\bf B}_j,\hat{\sigma}_{\epsilon_j}^2)}
	{{p}(y_{t+1}|\mathbf{x}_{t+1},  j,\mathcal{D}_{t})} = \mathcal{N}(\bbbeta_j; \hat{\bbmu}_{t+1}^j, \hat{\bbSig}^j_{t+1})\label{eq:theta_up}
\end{align}
where the updated mean $ \hat{\bbmu}_{t+1}^j$ and covariance matrix $\hat{\bbSig}^j_{t+1}$ are updated from slot to slot as in~\eqref{eq:beta_up}.

Please refer to Alg.~1 in the supplementary file for an overview of the ENS-LoRA-VBLL approach.

\section{Experimental results}
To test the efficacy of the proposed ENS-LoRA-VBLL, tests have been conducted on benchmark HDBO tasks as well as real-world molecular optimization tasks. Further, an extensive ablation study has been carried out.

The competing baselines consist of: i) GP~\cite{wan2021think}; ii) Infinite-width (I-)BNN~\cite{lee2017deep}; iii) Last Layer Laplace Approximation (LLLA)~\cite{kristiadi2024sober, mackay1992practical, immer2021improving}; iv) BLoB~\cite{wang2024blob}; v) Deep kernel learning (DKL)~\cite{wilson2016deep}; vi) Deep Ensembles (DE)~\cite{lakshminarayanan2017simple};  and vii) MLP-VBLL~\cite{brunzema2024bayesian}. 
For the fairness of comparison, we have replaced NNs in iii)-vi) by fine-tuned LLM with LoRA configuration~\cite{hu2022lora}, while maintaining GP and I-BNN implementations as in their original versions. All baselines adopt the TS~\cite{thompson1933likelihood, russo2018tutorial} as the AF. All the LLM-based baselines are implemented using the GPT-2 model~\cite{radford2019language} using the PEFT library~\cite{mangrulkar2022peft} for fine-tuning. We have also implemented LlamA-3.1-8B~\cite{grattafiori2024llama}, but the performance gain is marginal, while it is much more time-consuming.
Each experiment is repeated 5 times with different seeds. Due to space limitations, additional details about the experimental setup and more results are deferred to the supplementary file.


\subsection{Tests on benchmark HDBO tasks}





\begin{figure}
    \centering
    \includegraphics[width =\textwidth]{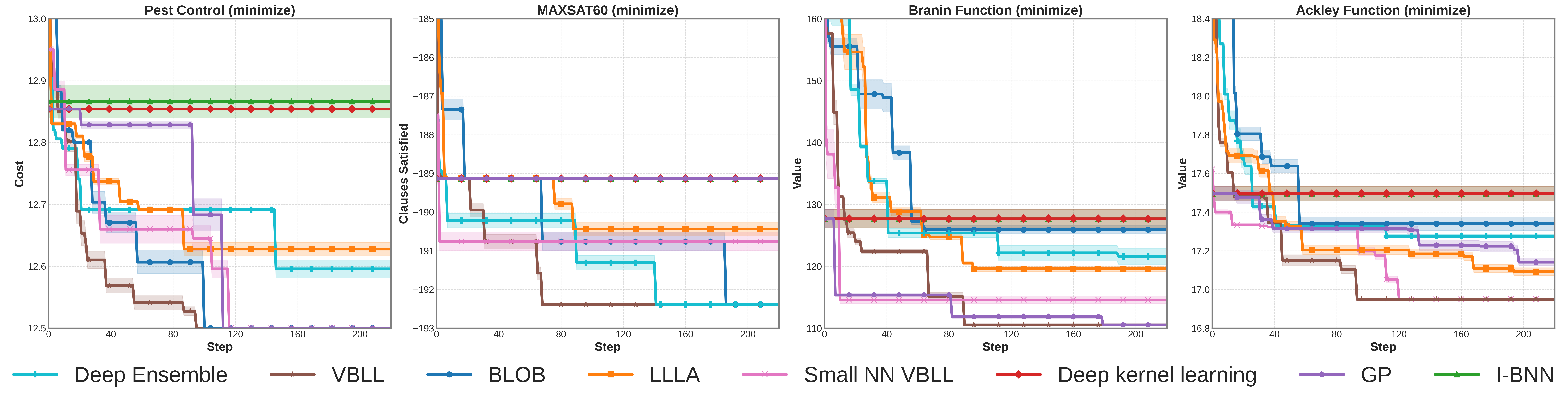}
        \vspace*{-0.5cm}
    \caption{Tests on benchmark HDBO tasks (note that these are minimization problems).}
    \label{fig: real_world optimization}  
\end{figure}


We first test on four standard synthetic and real HDBO tasks with different types of input variables, namely, \texttt{Branin}~\cite{branin1972widely}, \texttt{Ackley}~\cite{ackley2012connectionist}, \texttt{Pest Control}~\cite{gardner2014bayesian}, and \texttt{MAXSAT60}~\cite{bacchus2019maxsat}. The first two are synthetic functions with 32- and 60-dimensional continuous variables. \texttt{Pest Control} is a real-world categorical optimization task involving 25 categorical variables, each of which represents an intervention stage, taking one of five possible values\cite{oh2019combinatorial}. \texttt{MAXSAT60} represents a challenging Boolean satisfiability problem (60-dimensional binary optimization task), where, given a Boolean formula in conjunctive normal form, the objective is to find an assignment to the Boolean variables that maximizes the number of satisfied clauses~\cite{oh2019combinatorial}.

As shown in Fig.~\ref{fig: real_world optimization}, different approaches reveal distinct performance trends in various benchmark problems. In the \texttt{Pest Control} task, our ENS-LoRA-VBLL consistently achieves the lowest cost throughout the optimization process compared to other methods. Starting from similar initial cost values around 13.0, ENS-LoRA-VBLL rapidly decreases to approximately 12.63 by the 20th step, whereas LLLA and BLoB take over 40 steps to achieve comparable performance. By the 100th step, ENS-LoRA-VBLL reaches a cost of about 12.5, outperforming all other methods. Although MLP-VBLL eventually matches this performance, it requires nearly 120 iterations to do so.

\subsection{Tests on molecular optimization tasks}
 Molecular discovery represents a particularly challenging domain for BO due to its high dimensionality, inherent non-stationarity, and complex correlation structures. The search space of potential molecules is estimated to exceed $10^{100}$ unique structures \cite{restrepo2022chemical}, with only approximately $10^8$ experimentally synthesized compounds reported to date. In this setting, effective surrogate models must capture meaningful patterns from limited data while providing well-calibrated uncertainty estimates to guide the exploration of this vast chemical space. Four chemical optimization tasks selected from recent literature: (i) \texttt{redoxmer} - redox potential minimization~\cite{agarwal2021discovery}, (ii) \texttt{solvation} - solvation energy minimization~\cite{agarwal2021discovery}, (iii) \texttt{Kinase} - docking score minimization for kinase inhibitors ~\cite{graff2021accelerating}, and (iv) \texttt{Laser} -fluorescence oscillator strength maximization for laser materials \cite{strieth2024delocalized}. 
Following \cite{kristiadi2024sober}, we use SMILES~\cite{weininger1988smiles} representations as inputs for our surrogate models. For models requiring vector inputs (GP, I-BNN), we employ Morgan fingerprints~\cite{morgan1965generation} as the molecular featurization method. 


\begin{figure}
    \centering
    \includegraphics[width = \textwidth]{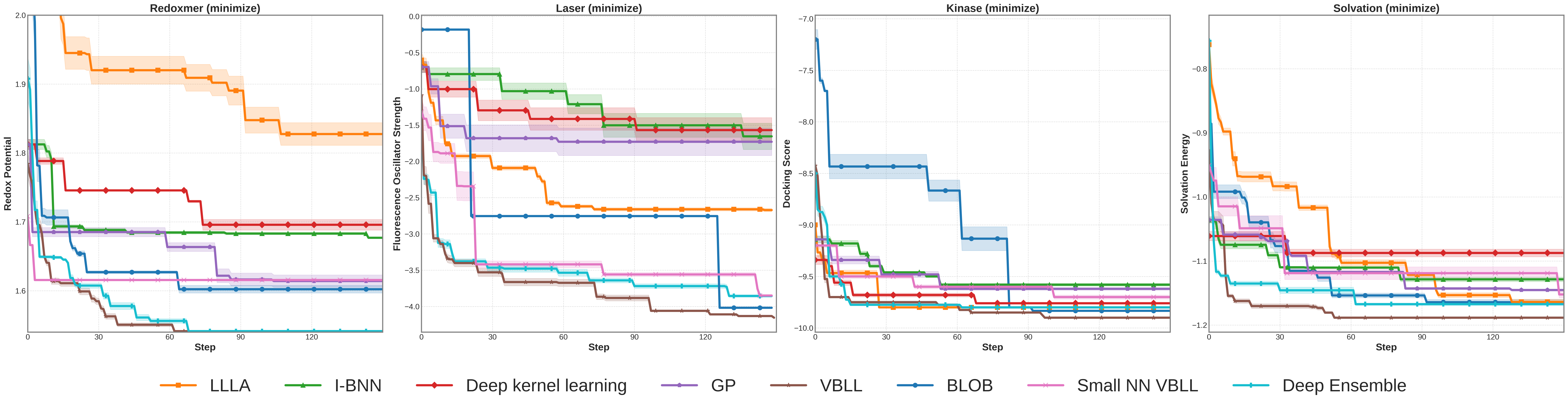}
    \vspace*{-0.6cm}
    \caption{Tests on different molecular optimization tasks.}
    \label{fig: molecular optimization}  
\end{figure}

As shown in~Fig.~\ref{fig: molecular optimization}, the molecular optimization study indicates that our ENS-LoRA-VBLL performs effectively in ainr tasks. In the \texttt{Redoxmer} task, both the ENS-LoRA-VBLL and DE methods rapidly converge to lower values, surpassing the performance of LLLA and I-BNN. For the \texttt{Laser} task, MLP-VBLL and our LLM-based ENS-LoRA-VBLL approach achieve better results than Gaussian Process (GP) methods. In the \texttt{Kinase} inhibitor experiment, the ENS-LoRA-VBLL method maintains a steady optimization trajectory, while GP methods plateau after showing initial promise. Overall, our ENS-LoRA-VBLL method offers competitive optimization performance with lower computational demands compared to BLoB, especially in HD discrete optimization scenarios.

\subsection{Ablation study}
\subsubsection{Effect of fine-tuning and pre-training}
Here, we compared on the \texttt{Redoxmer} task (i) our~\textbf{ENS-LoRA-VBLL} with LLM backbone with fine-tuned LoRA parameters; (ii) \textbf{MLP-VBLL} with MLP backbone; (iii) \textbf{LLM-BLL} using pre-trained LLM-based feature extractor; and (iv) \textbf{Rand-LLM-BLL} using randomly initialized LLM without pre-training.

It is evident from Fig.~\ref{fig: ablation} (a) 
that our ENS-LoRA-VBLL approach consistently outperforms all variants, achieving the lowest redox potential. The Rand-LLM-BLL variant's poor performance confirms the value of LLM pre-training as a molecular representation prior, while MLP-VBLL's limitations demonstrate the benefits of larger model capacity. These results validate that the combining method enhances sample efficiency in HDBO

\subsubsection{Fine-tuning and recursive update}

We evaluated three update strategies in our ENS-LoRA-VBLL framework on the \texttt{Redoxmer} task, namely, \textbf{fine-tuning} in every iteration, \textbf{event-triggered updates}, and \textbf{recursive updates} without fine-tuning. The results according to the Fig.~\ref{fig: ablation} (b) reveal a clear efficiency-performance tradeoff. Full fine-tuning delivers strong results but with the highest computational cost. Recursive updates offer the most efficient approach while achieving comparable results after few more steps. Event-triggered updates provide a balanced middle ground, selectively invoking fine-tuning only when necessary (i.e. below predictive likelihood).
\vspace{-2mm}
\subsubsection{Effect of ensembling ranks and weighting strategy}
\begin{figure}
    \centering
\includegraphics[width =\textwidth]{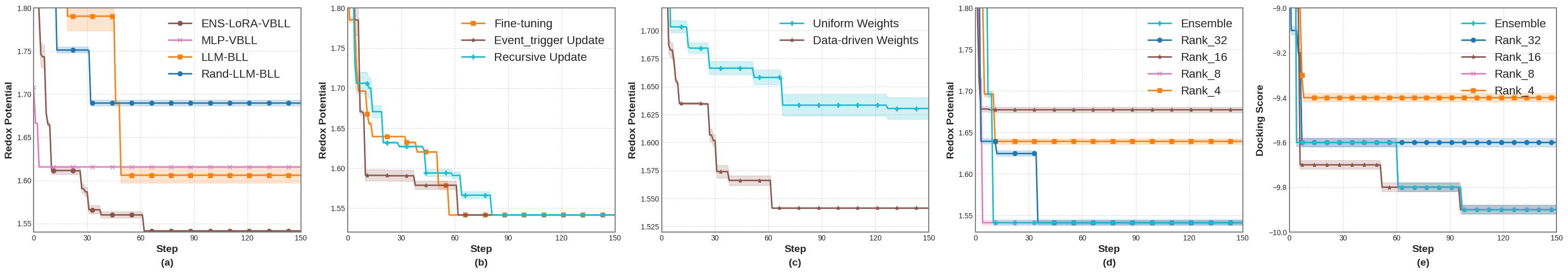}
        \vspace*{-0.5cm}
        \caption{Ablation study}
    \label{fig: ablation}  
\end{figure}

Fig.\ref{fig: ablation} (c) demonstrates that our ENS-LoRA-VBLL with data-driven weights~\eqref{eq: weight} consistently outperforms the uniform weighting approach across all optimization stages for \texttt{redoxmer} minimization task, achieving a lower final redox potential.

 Fig.~\ref{fig: ablation} (d-e) shows that different LoRA ranks exhibit varying effectiveness across datasets: in \texttt{redoxmer} minimization, LoRA-VBLL with rank 8 achieves optimal performance, while in \texttt{kinase} optimization, LoRA-VBLL with rank 16 performs the best. By contrast,the  ENS-LoRA-VBLL approach consistently delivers superior results in both tasks by effectively combining strengths from multiple ranks, which aligns with the idea of~\cite{chronopoulou2023adaptersoup}. This demonstrates ENS-LoRA-VBLL's ability to adaptively leverage the most suitable rank for each specific molecular optimization problem, providing robust performance across diverse chemical spaces without requiring manual rank selection.
\vspace{-0.7mm}


\vspace{-1.0mm}


\vspace{-4mm}


\begin{figure}
    \centering
    \includegraphics[width=\linewidth]{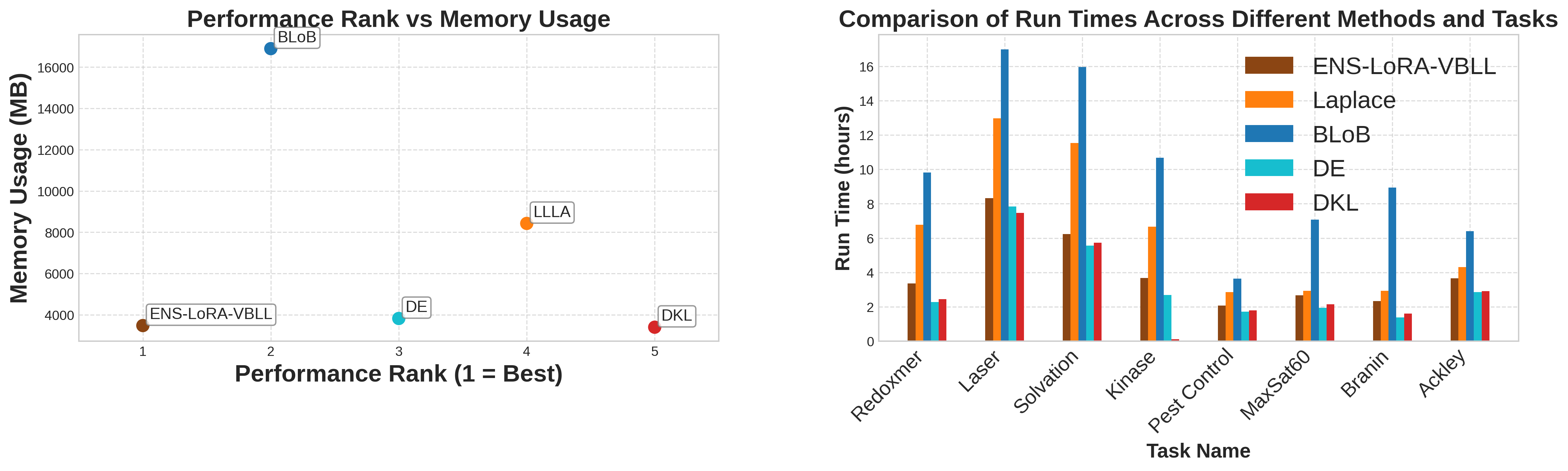}
        \vspace*{-0.7cm}
    \caption{Comparison of memory and runtime}
    \label{fig: memory and runtime}
\end{figure}

\subsection{Comparison of memory and runtime}

Here, we compare all LLM-based methods using identical GPT-2 backbones, full-precision, batch sizes, and tokenizer settings, with LoRA configurations evaluated across ranks {4, 8, 16, 32}. Results associated with each method's best-performing configuration are reported. 
As shown in Fig.~\ref{fig: memory and runtime}, our ENS-LoRA-VBLL approach achieves an optimal performance-efficiency balance. It requires only 3.5GB of memory (78\% less than BLoB's 16.8GB) while delivering top performance. BLoB's memory demands stem from Monte Carlo sampling, while LLLA (8.4GB) performs worse due to post-hoc uncertainty approximation. Small NN VBLL uses minimal memory (0.7GB) but with reduced effectiveness. ENS-LoRA-VBLL's efficiency comes from analytical ELBO computation, event-triggered recursive updates, and parameter-efficient adaptation across all tested ranks.




\section{Conclusion and discussion}
This paper built on the VBLL framework to put forth a novel LLM-based surrogate with LoRA-based parameter-efficient fine-tuning for HDBO with irregular variables. The LoRA parameters are jointly optimized with the variational posterior. Between any two fine-tuning steps, the resultant LoRA-VBLL framework affords recursive update with scalability. To further bypass the need to manually select the LoRA, we employ an ensemble of LoRA-VBLL models, each of which has a scalable recursive update in parallel. The per-model data-adaptive weight in ENS-LoRA-VBLL also admits a closed-form recursive update. Extensive results on both synthetic and real HD tasks corroborate the consistently superior performance of the (ENS)-LoRA-VBLL approaches relative to GP- and LLM-based baselines.

\noindent{\bf Limitations.} This work adapts LLMs for HDBO tasks. The AF design follows the current practise. However, each evaluation of the HD AF could still be time-consuming, given the need to pass through the LLM-based feature extractor. Although caching the features can reduce the computational complexity if ${\bf x}$ is chosen from a pool of candidates (as in the molecular optimization task), further improvement is due for the general case that entails gradient evaluation for continuous variables and local search for categorical variables.

\noindent{\bf Societal impact.} Given the ubiquity of HD black-box optimization tasks in the wild (e.g., drug discovery, robotics, molecular optimization, etc.), this work will have substantial societal impact. 
\clearpage
\appendix

\section{Converting BO task to language task}
To allow for BO tasks to be handled by LLMs, we need to convert structured data into natural language prompts. This conversion enables LLMs to serve as surrogate models without architectural modifications. Unlike traditional surrogate models such as GPs or BNNs, which require specialized implementations to handle different input spaces, LLMs can process various types of structured data through the same text interface.

The language-based conversion process serves multiple purposes in our optimization framework. First, it allows us to maintain the original LLM architecture and training objective, eliminating the need for customized output layers or loss functions. Second, it provides a unified interface for handling diverse optimization problems spanning different domains, variable types, and dimensionalities. Third, it leverages the semantic understanding capabilities of pretrained LLMs, potentially improving their ability to capture complex relationships between optimization variables and the objective values.

Take the \textbf{MAXSAT-60 Problem} for example.
This problem involves finding Boolean assignments to 60 variables that maximize formula satisfaction. Consider this example from our dataset:

\begin{table}[h]
\centering
\caption{MAXSAT-60 vs Objective Value}
\begin{tabular}{cc@{}c@{}ccc|c}
\hline
$x_0$ & $x_1$ & \multicolumn{3}{c}{$\cdots$} & $x_{59}$ & $MAXSAT60$ \\
\hline
0 & 0 & \multicolumn{3}{c}{$\cdots$} & 0 & $-139.57378$ \\
\hline
\end{tabular}
\end{table}

This tabular data can be converted into a natural language prompt following LIFT\cite{dinh2022lift}  with a specific format for training:
\begin{verbatim}
"For a MAXSAT-60 problem with Boolean assignment x_0=0, x_1=0, and x_59=0,
the objective value is: -139.57378"
\end{verbatim}
For inference with a new configuration, we create a prompt with a blank space for the model to complete:
\begin{verbatim}
"For a MAXSAT-60 problem with Boolean assignment x_0=1, x_1=0, and x_59=1,
the objective value is: "
\end{verbatim}

By explicitly instructing the model to "respond with only a number," we constrain the output format to be a numerical value without additional explanatory text. This approach simplifies post-processing and ensures the model generates predictions in a format that can be directly used by the AF in the BO loop. This explicit instruction helps guide the LLM to produce clean, numerical outputs that are easy to parse and use in subsequent optimization steps.

Our prompt template uses color-coding to highlight different components:

\begin{itemize}
    \item \textcolor{orange}{Orange text} - Task information: optimization objective, task name, and metric
    \item \textcolor{yellow}{Yellow text} - Problem features: description of the input space and variables
    \item \textcolor{blue}{Blue text} - Observations: previous evaluations and new configurations to analyze
    \item \textcolor{green}{Green text} - Response instructions: format guidelines for the model output
\end{itemize}

\begin{tcolorbox}[
  colback=gray!5,
  colframe=gray!40,
  title=General Template for Language-Based BO Prompts,
  listing only,
  listing options={
    basicstyle=\ttfamily,
    breaklines=true,
    escapeinside={(*@}{@*)}
  }
]
[INST] You are assisting me with automated machine learning using (*@\textcolor{orange}{optimization\_task}@*). The (*@\textcolor{orange}{task\_name}@*) performance is measured using (*@\textcolor{orange}{metric}@*).

The problem involves (*@\textcolor{yellow}{feature\_description}@*).

I have collected the following observations: (*@\textcolor{blue}{previous\_observations}@*)

Based on these observations, what would be the (*@\textcolor{orange}{metric}@*) for the following configuration? (*@\textcolor{blue}{new\_configuration}@*)

(*@\textcolor{green}{Please respond with only a numerical value.}@*) [/INST]
\end{tcolorbox}


\section{Comparison of Bayesian fine-tuning frameworks for LLMs}
\begin{figure}
    \centering
    \includegraphics[width=0.8\linewidth]{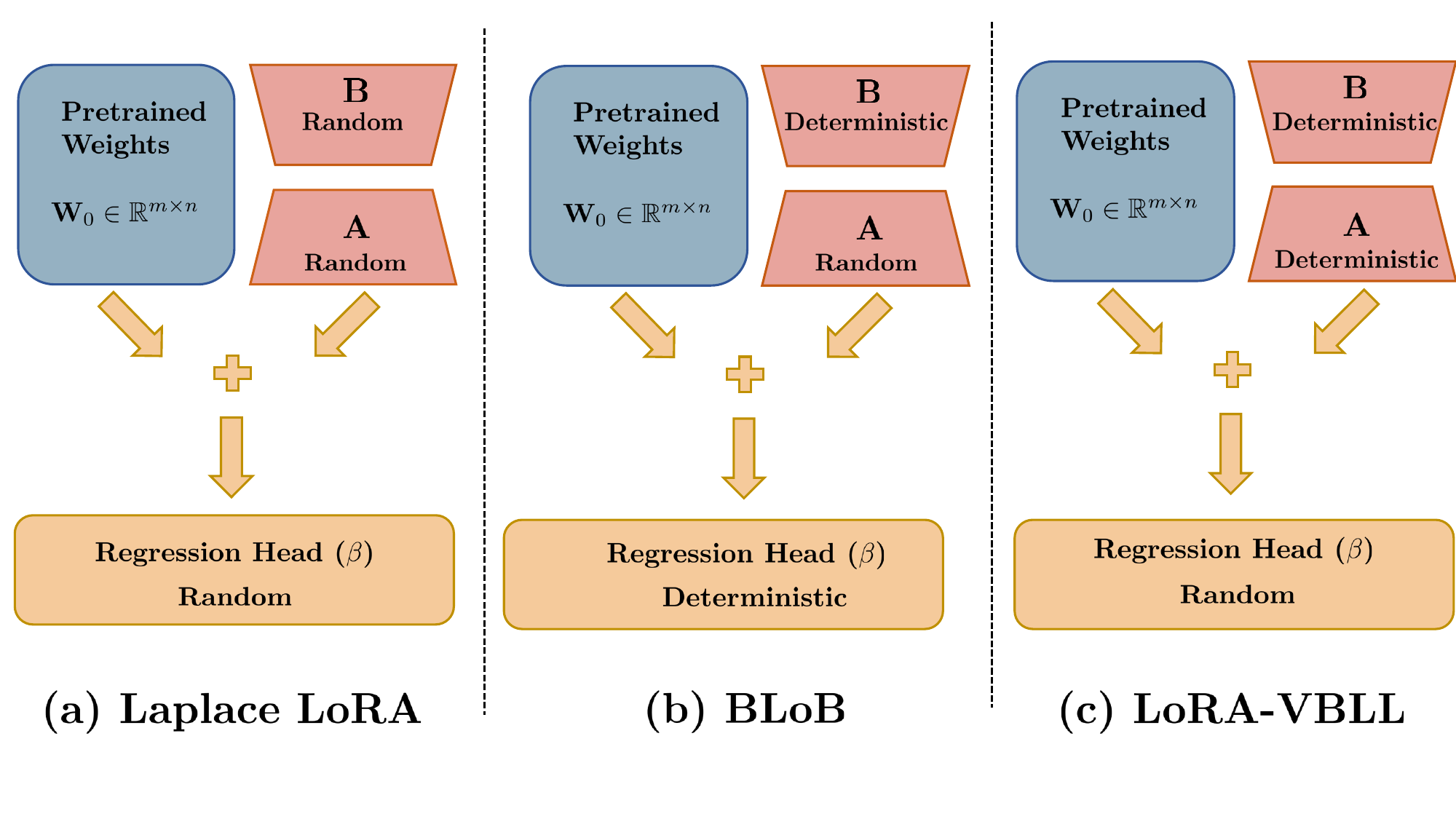}
    \caption{Architectures for adapting LLMs as surrogate models: (a) Laplace LoRA, (b) BLoB, and (c) LoRA-VBLL ({\bf proposed})}
    \label{fig:variants for LLM-based surrogate models}
\end{figure}

The three approaches illustrated in Figure~\ref{fig:variants for LLM-based surrogate models} represent different strategies for adapting pretrained LLMs as surrogate models through varying inference techniques.

\textbf{Laplace LoRA} (Figure~\ref{fig:variants for LLM-based surrogate models}(a)) performs post-hoc Laplace approximation on both the LoRA adapters (${\bf A}$ and ${\bf B}$) and the regression head ($\bbbeta$), providing uncertainty estimates for all trainable parameters after standard regression training~\cite{kristiadi2024sober, yang2023bayesian}.

\textbf{BLoB} (Figure~\ref{fig:variants for LLM-based surrogate models}(b)) employs Bayes by Backprop variational inference specifically for the adapter ${\bf A}$, enabling stochastic gradient-based learning of the posterior distribution during training, while maintaining deterministic initialization for the regression head ~\cite{wang2024blob}.

\textbf{LoRA-VBLL} (Figure~\ref{fig:variants for LLM-based surrogate models}(c)) relies on the Bayesian laster layer model with {\it determinstic} LLM-based feature extractor and {\it random} regression head. This architecture is amenable to scalable variational training with closed-form recursive posterior updates.  The deterministic initialization of adapters ensures stable feature extraction from the pretrained LLM, while the Bayesian treatment of the regression layer captures predictive uncertainty crucial for effective BO in high-dimensional spaces.

\section{Algorithm}
Please refer to Algs.1-2 for the detailed implementation of the proposed ENS-LoRA-VBLL approach.

\begin{algorithm}[t]
\caption{ENS-LoRA-VBLL}
\label{alg:ENS-LoRA-VBLL}
\begin{algorithmic}[1]
\STATE{\bf Input:} Initial data $\mathcal{D}_0 := \{({\bf x}_\tau, y_\tau)\}_{\tau=1}^{n_0}$, pre-trained LLM weights ${\bf W}_0$,  performance threshold $\gamma$, decay factor $\alpha$, evaluation budget $T$, number of ranks $J$, ranks $\{r_j\}_{j=1}^J$, prior variances $\{\sigma_{\beta_j}^2\}_{j=1}^J$, prior weight $w_0^j = 1/J$, $j = 1, \ldots, J$
\newline
\FOR{$j = 1$ to $J$}
    \STATE Obtain $\hat{\boldsymbol{\Theta}}_j$ and $\mathcal{L}_0^{\text{ELBO}}(\hat{\boldsymbol{\Theta}}_j)$ via Alg.~2;
    \STATE Update $w_0^j$ using $\mathcal{L}_0^{\text{ELBO}}(\hat{\boldsymbol{\Theta}}_j)$ according to ~\eqref{eq: weight} and~\eqref{eq: weight_ELBO};
\ENDFOR
\newline
\FOR{$t = 0$ to $T-1$}
    \STATE Obtain ${\bf x}_{t+1}$ by optimizing the AF~\eqref{eq:ensemble_AF};
    \STATE Evaluate ${\bf x}_{t+1}$ to obtain $y_{t+1}$;
    \STATE Augment data $\mathcal{D}_{t+1} := \mathcal{D}_t \cup \{({\bf x}_{t+1}, y_{t+1})\}$;
    \STATE Calculate the predictive likelihood $p(y_{t+1}|{\bf x}_{t+1}, {\cal D}_t)$;
\newline
    \IF{$p(y_{t+1}|{\bf x}_{t+1}, {\cal D}_t)<\gamma$}
    \FOR{$j = 1$ to $J$}
    \STATE Perform fine-tuning to obtain updated $\hat{\boldsymbol{\Theta}}_j$ using Alg.~2 and ${\cal D}_{t+1}$;
        \STATE Obtain $w_t^j$ using $\mathcal{L}_t^{\text{ELBO}}(\hat{\boldsymbol{\Theta}}_j)$ according to ~\eqref{eq: weight} and~\eqref{eq: weight_ELBO};
   \ENDFOR
   \ELSE
    \FOR{$j = 1$ to $J$}
     \STATE Recursive updates of $\{w_t^j, \hat{\bbmu}_t^j, \hat{\bbSig}_t^j, j\in\mathcal{J}\}$ via ~\eqref{eq:w_update}-\eqref{eq:theta_up};
    \ENDFOR
     \ENDIF   
     \newline
\ENDFOR
\newline
\STATE \textbf{Output:} ${\bf x}^* = {\bf x}_{\tau^*}$, where $\tau^* = \underset{\tau \in \{1,\ldots, n_0+T \}}{\arg\max} \ \  y_\tau$
\end{algorithmic}
\end{algorithm}

\begin{algorithm}[t]
\caption{Training a single LoRA-VBLL via the ELBO objective}
\label{alg:tain_vbll_lora}
\begin{algorithmic}[1]
\REQUIRE Data $\mathcal{D}_t := \{({\bf x}_\tau, y_\tau)\}_{\tau=1}^{t}$, pre-trained LLM weights ${\bf W}_0$, rank $r$, prior variance $\sigma_\beta^2$, initial value for $\sigma_\epsilon^2$, learning rate $\eta$, number of iterations $I$
\newline
\STATE Initialize LoRA parameters ${\bf A} \in \mathbb{R}^{r \times m}$, ${\bf B} \in \mathbb{R}^{r \times n}$ randomly
\STATE Initialize posterior $q_t(\bbbeta) = \mathcal{N}(\bbbeta; \boldsymbol{\mu}_t, \boldsymbol{\Sigma}_t)$ where $\boldsymbol{\mu}_t = \mathbf{0}$, $\boldsymbol{\Sigma}_t = \sigma_\beta^2{\bf I}$

\FOR{$i = 1$ to $I$}
    \STATE Sample mini-batch $\mathcal{B}_i \subset \mathcal{D}_t$
    \STATE Compute feature mapping $\bbphi_{{\bf W}_0+{\bf A}^\top {\bf B}}({\bf x})$ for all ${\bf x} \in \mathcal{B}_j$
    \STATE Compute the gradient of the ELBO~\eqref{eq:ELBO}
       $\nabla_{\boldsymbol{\Theta}} \mathcal{L}_t^{\text{ELBO}}(\boldsymbol{\Theta})$
    \STATE Update parameters:
     $ \hat{\boldsymbol{\Theta}} \gets \hat{\boldsymbol{\Theta}} + \eta \cdot \nabla_{\boldsymbol{\Theta}} \mathcal{L}_t^{\text{ELBO}}$
\ENDFOR
\newline
\STATE \textbf{Output:} $\hat{\boldsymbol{\Theta}}$ and $\mathcal{L}_t^{\text{ELBO}}(\hat{\boldsymbol{\Theta}})$
\end{algorithmic}
\end{algorithm}

\section{Additional experimental results}

\subsection{Benchmark HDBO Tasks}
\label{appendix:benchmark_tasks}

In our experiments, we evaluate our proposed approaches on several benchmark high-dimensional Bayesian optimization (HDBO) tasks with different types of variables:

\subsubsection{Categorical Variable Problems}

\begin{itemize}
    \item \textbf{Branin32}~\cite{branin1972widely}: A standard test function extended to 32 dimensions with categorical variables. Each dimension is discretized into categorical levels, creating a challenging discrete optimization problem with multiple local minima. The underlying continuous function structure is preserved through discretization:
    
    \begin{equation}
        f(x_1, x_2) = (x_2 - 0.129x_1^2 + 1.6x_1 - 6)^2 + 10(1 - 0.125/\pi)\cos(x_1) + 10
    \end{equation}
    
    where the continuous variables are mapped to discrete categorical values across all 32 dimensions.
    
    \item \textbf{Ackley60}~\cite{ackley2012connectionist}: A multimodal test function discretized into 60 categorical dimensions. The discrete version maintains the challenging landscape of numerous local minima while operating over categorical variables. The original continuous formulation:
        
    \begin{equation}
        f(x) = -20\exp\left(-0.2\sqrt{\frac{1}{d}\sum_{i=1}^{d}x_i^2}\right) - \exp\left(\frac{1}{d}\sum_{i=1}^{d}\cos(2\pi x_i)\right) + 20 + e
    \end{equation}
    
    is evaluated over discretized categorical inputs, where each dimension takes values from a finite categorical set.
    
    \item \textbf{Pest Control}~\cite{gardner2014bayesian}: A real-world problem involving 25 categorical variables, each representing an intervention stage with five possible values. Each configuration in this problem defines a pest control strategy, and the objective is to minimize the cost associated with pest population dynamics. This benchmark represents a challenging categorical optimization task with real-world relevance in ecological management.

    \item \textbf{MAXSAT60}~\cite{bacchus2019maxsat}: A Boolean satisfiability problem with 60 binary variables. The objective is to find an assignment to these variables that maximizes the number of satisfied clauses in a given Boolean formula. MAXSAT60 represents a purely discrete optimization challenge that is NP-hard in nature, testing our method's ability to navigate combinatorial search spaces~\cite{oh2019combinatorial}.
    
\end{itemize}

\subsubsection{Mixed and Continuous Variable Problems}

\begin{itemize}
    \item \textbf{Ackley53}~\cite{kandasamy2018parallelised}: A modified version of the Ackley function that incorporates mixed variable types. This benchmark includes 53 dimensions with 3 continuous variables and 50 categorical variables. The categorical variables are one-hot encoded, resulting in a challenging mixed-variable optimization problem that tests our method's ability to handle heterogeneous input spaces.

    \item \textbf{Ackley30}~\cite{kandasamy2018parallelised}: A modified version of the Ackley function that incorporates continuous variable types. This benchmark includes 30 dimensions with both continuous variables. It is a challenging, continuous optimization problem.

\end{itemize}

In practice, for computational efficiency under continuous variables, we pre-generate a discrete candidate set using Sobol sequences \cite{eriksson2019scalable, lin2023use}, allowing AF optimization to proceed via simple enumeration rather than expensive gradient-based continuous optimization.

\subsection{Molecular Optimization Tasks}
\label{appendix:molecular_tasks}

We also evaluate our approach on molecular optimization tasks, which involve complex, high-dimensional search spaces with discrete structures:

\begin{itemize}
    \item \textbf{Redoxmer}~\cite{agarwal2021discovery}: This task focuses on minimizing the redox potential of molecules for possible flow battery electrolytes. The search space consists of molecules represented in SMILES format, with the objective function computed using physics-inspired simulators.

    \item \textbf{Solvation}~\cite{agarwal2021discovery}: Similar to Redoxmer, this task aims to minimize the solvation energy of molecules. Solvation energy is a critical property for flow battery electrolytes and influences their stability and performance.

    \item \textbf{Kinase}~\cite{graff2021accelerating}: This task involves minimizing the docking score of kinase inhibitors for drug discovery applications. The objective function simulates the binding affinity of molecules to target kinase proteins.

    \item \textbf{Laser}~\cite{strieth2024delocalized}: This task aims to maximize the fluorescence oscillator strength of molecules for laser applications. The search space consists of potential laser materials represented as molecular structures.
\end{itemize}

Each molecular optimization task represents a real-world challenge where efficient exploration of the chemical space is crucial. These problems deal with high-dimensional, discrete search spaces with complex structural dependencies that standard optimization methods struggle to handle effectively.

\subsection{Experiment Setup}

We compare the performance of the following surrogate models. 

\paragraph{GP:} As the de-facto standard in BO, we use GPs with a Matérn $5/2$ kernel ($\nu = 2.5$) and individual lengthscales $l_i$ for all input dimensions. These lengthscales are optimized within box constraints $l_i \in [0.005, 4]$ following recommended best practices \cite{eriksson2019scalable, balandat2020botorch}. For single-objective problems, we use an adaptive scale with a lengthscale prior of $\mathrm{Gamma}(3, 6)$ and an output-scale prior of $\mathrm{Gamma}(2.0, 0.15)$, which combine with the marginal likelihood to form a posterior that we optimize for hyperparameter learning. 

\paragraph{Infinite-width (I-) BNN:} Following the setting in~\cite{li2024study}, we employ I-BNNs\cite{lee2017deep} with $3$ hidden layers and ReLU activation functions. We set the weight variance to $10.0$ and the bias variance to $1.6$. Note that the I-BNN is expressed as a kernel; therefore, the model is still non-parametric and does not learn features. 

For a fair comparison in our LLM-based Bayesian optimization framework, we have replaced all neural network components in the following methods with LLM\cite{brown2020language}+LoRA\cite{hu2022lora} configurations, while maintaining GP and I-BNN implementations of these two non-parametric methods in their original versions.

\paragraph{Deep Kernel Learning (DKL):} DKL \cite{wilson2016deep} combines feature extraction with neural networks and GPs. In our implementation, we adapt this approach to utilize LLMs as feature extractors. Specifically, we fine-tune LlaMA3.1-8B\cite{dubey2024llama} using LoRA on the output layer and the queries and values of all attention layers. This LLM-based feature extractor $g_\theta$ is then wrapped in a kernel as $k_{\text{DKL}}:= k(g_\theta(x), g_\theta(x'))$, allowing for non-stationary modeling and exact inference. We set up the base kernel using the same Matérn $5/2$ kernel as in our GP implementation. For multi-objective problems, we independently model each objective. 


\paragraph{Last Layer Laplace Approximation (LLLA):} LLLA provides a computationally efficient way to obtain uncertainty estimates after training a neural network \cite{mackay1992practical, immer2021improving}. In our approach, we apply LLLA to fine-tuned LlaMA3.1-8B models with LoRA adaptations. Following Laplace-LoRA \cite{kristiadi2024sober, yang2023bayesian}, we apply the Laplace approximation to the LoRA parameters in the output layer and attention layers to obtain a Gaussian posterior over these parameters. 

\paragraph{Deep Ensembles (DE):} Deep ensembles\cite{lakshminarayanan2017simple} offer a computationally efficient approach to obtaining predictions with uncertainty. Instead of traditional neural networks, we implement an ensemble of $4$ LlaMA3.1-8B models, each fine-tuned with different LoRA parameters. Each model uses the same base LLM architecture but has independently trained LoRA parameters, effectively creating diversity in the ensemble while maintaining computational efficiency. Each model is trained on a random 80\% of the function evaluations. We parameterize the predictive as a normal distribution with mean and variance derived from the ensemble. 

\paragraph{Bayesian Low-Rank Adaptation (BLoB):} BLoB \cite{wang2024blob} provides a principled Bayesian framework for LLM-based surrogate models. We implement BLoB using LlaMA3.1-8B with LoRA applied to both the output layer and the queries and values of all attention layers. Unlike post-training Bayesian methods, BLoB jointly adjusts both the mean and covariance of the LLM parameters throughout the fine-tuning process, leading to better calibrated uncertainty estimates. We maintain the default hyperparameter settings from the original implementation for reproducibility, including the LoRA rank and learning rate. 

\paragraph{MLP-VBLL:} To illustrate the power of large models, we choose as baseline VBLL with an MLP with $3$ hidden layers and ReLU activation functions following the original work that adapts VBLL for BO~\cite{harrison2024variational}.

\subsection{Results using different AFs}

\begin{figure}
    \centering
    \includegraphics[width=0.9\linewidth]{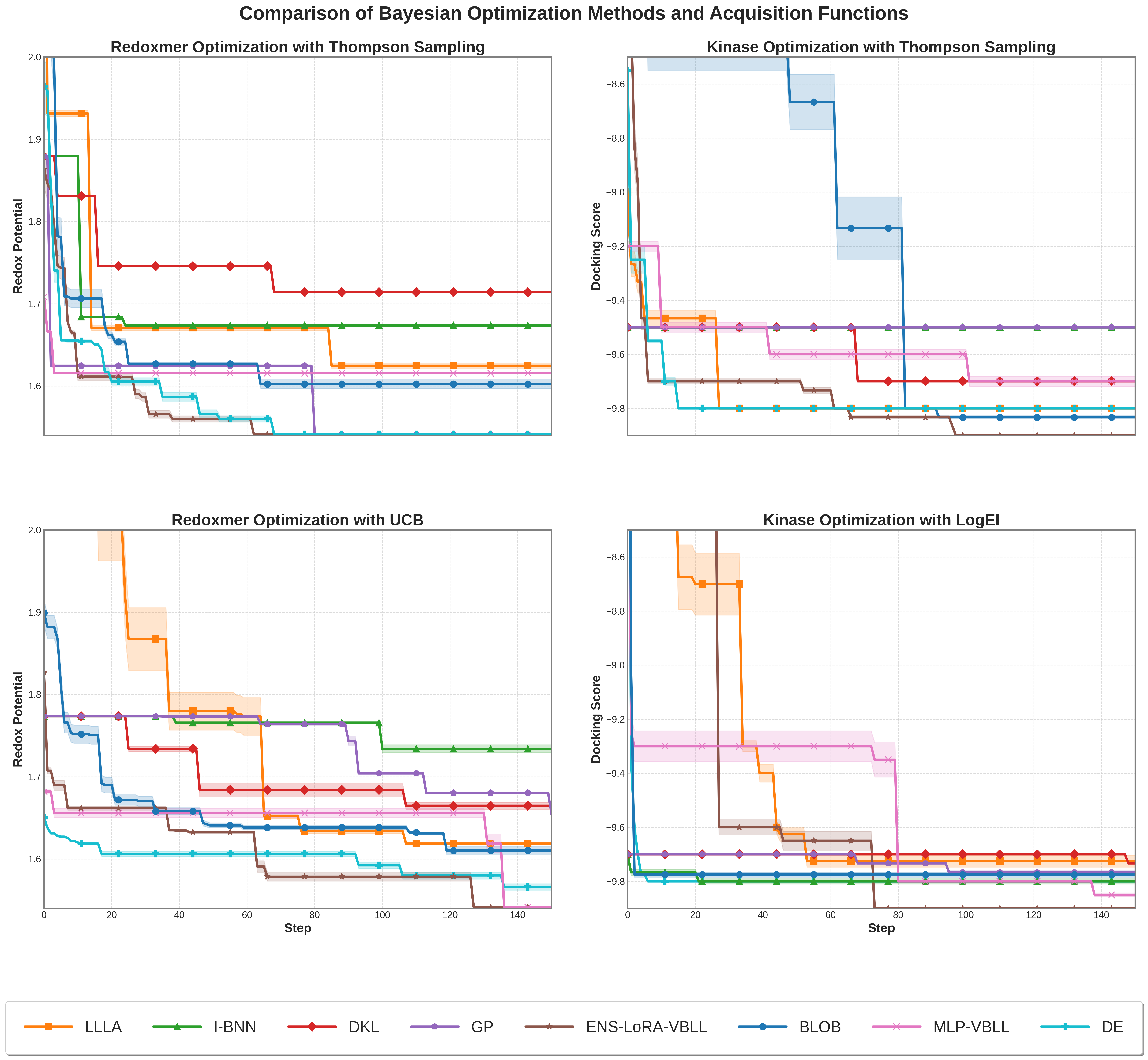}
    \caption{Performance on Molecular Optimization taks using different AFs}
    \label{fig:comparison}
\end{figure}

To assess the robustness of ENS-LoRA-VBLL across acquisition strategies, we evaluated Thompson Sampling (TS), Upper Confidence Bound (UCB), and LogEI on molecular optimization tasks. As shown in Figure~\ref{fig:comparison}, the choice of acquisition function significantly impacts baseline methods: LLLA and DE exhibit substantial performance variations across different AFs, with DE showing particularly high variance under UCB. In contrast, ENS-LoRA-VBLL maintains exceptional consistency and superior performance regardless of the acquisition strategy, demonstrating that our LLM-based surrogate model provides robust uncertainty estimates that work effectively across different exploration-exploitation trade-offs. This AF-agnostic robustness eliminates the need for careful acquisition function tuning and reinforces the practical applicability of our approach.

\subsection{HDBO over continuous and mixed variables}

\begin{figure}
    \centering
    \includegraphics[width=0.9\linewidth]{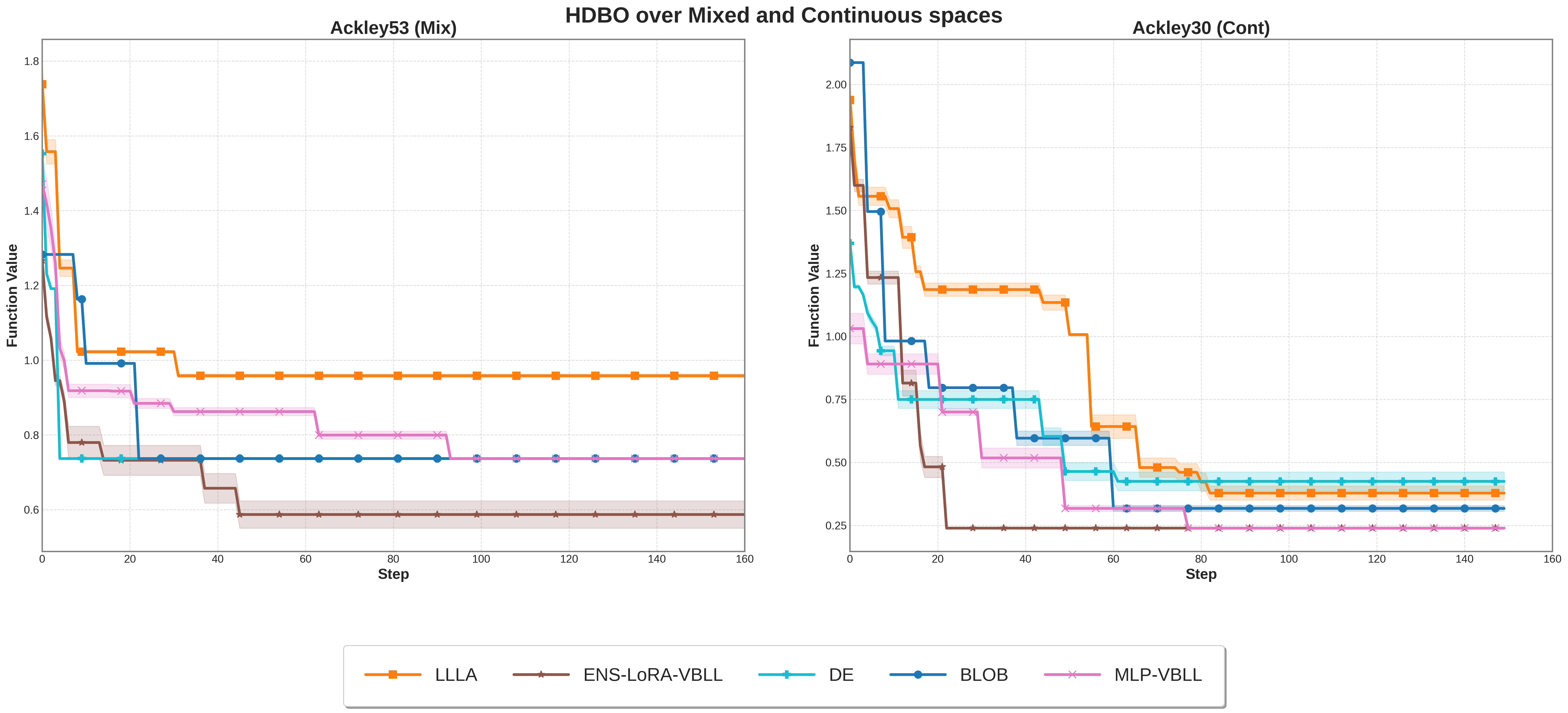}
    \caption{HDBO over Mixed and Continuous spaces}
    \label{fig:HDBO over mixed/continuous/ spaces}
\end{figure}

To evaluate our method's versatility across different variable types, we conducted experiments on both mixed-variable and continuous optimization problems. According to the fig\ref{fig:HDBO over mixed/continuous/ spaces}, for the continuous Ackley30 benchmark, ENS-LoRA-VBLL shows competitive performance, closely matching BLoB while significantly outperforming LLLA and MLP-VBLL. The continuous space results reveal distinct convergence patterns: LLLA shows initial promise but plateaus early, while BLoB exhibits more consistent performance in continuous spaces compared to mixed-variable settings. 

Our method maintains robust performance across both variable types, with particularly strong advantages in mixed-variable scenarios where the discrete-continuous interface poses additional modeling challenges. These results confirm ENS-LoRA-VBLL's effectiveness in handling heterogeneous optimization spaces and demonstrate the practical value of LLM-based feature extraction for diverse optimization problems.

\section{Implementation Details}

Our implementation is built using PyTorch\footnote{https://pytorch.org/}\cite{paszke2019pytorch} with the Transformers library\footnote{https://huggingface.co/docs/transformers/en/index}\cite{wolf2019huggingface} for LLM backbone integration. We utilize the PEFT library \cite{mangrulkar2022peft} for parameter-efficient fine-tuning with LoRA configurations. The posterior $q_t(\bbbeta)$ in VBLL is implemented with custom multivariate normal distributions supporting multiple parameterizations, including dense, diagonal, low-rank, and dense precision formulations.

For efficient uncertainty quantification, we implement custom Cholesky update operations using rank-1 updates for recursive Bayesian inference. The dense precision parameterization utilizes specialized matrix operations including Cholesky decomposition, matrix inversion via \texttt{torch.cholesky\_solve}, and symmetric matrix operations. Our recursive update mechanism leverages batched tensor operations to maintain computational efficiency during the optimization loop.

The proposed ENS-LoRA-VBLL combines a fine-tuned LLM through LoRA with different ranks from $\{8, 16, 32, 64\}$ through adaptive weighting based on marginal likelihood estimation. Feature caching is implemented for the molecular optimization tasks to avoid redundant forward passes through the LLM backbone between fine-tuning steps, significantly reducing computational overhead.

\subsection{Training Procedure}

Our training procedure consists of two distinct phases designed to ensure stable convergence and optimal performance. In the first phase, we perform MSE-based initialization where only the mean parameters ($\mathbf{W}_{\text{mean}}$) and noise variance are optimized using standard mean squared error loss for 500 epochs. This provides a stable initialization point for the subsequent variational training phase.

In the second phase, we conduct full ELBO-based training with all VBLL parameters using AdamW optimizer~\cite{loshchilov2017decoupled} with learning rate $1 \times 10^{-3}$ and weight decay $1 \times 10^{-4}$. 

For the extra VBLL hyperparameters used in the original codebase of VBLL\footnote{https://github.com/VectorInstitute/vbll/tree/main}. We use the default configuration. The prior scale is set to $10.0$ and the Wishart scale to $1 \times 10^{-3}$ with the degrees of freedom set to $2.0$. 

LoRA configurations use the $\alpha$ parameter of 32 and a dropout rate of 0.1. The ensemble weights are determined using marginal likelihood estimation with a temperature scaling parameter of 1.0. 

\subsection{ Feature caching for efficient recursive updates.}

The computational bottleneck in the recursive Bayesian update of Eq.~\eqref{eq:theta_up} lies in the repeated forward passes through the LLM to compute the feature mapping $\bbphi_{{\bf W}_0+{\bf A}^\top {\bf B}}({\bf x})$. Between two consecutive fine-tuning steps where the LoRA parameters $\{{\bf A}, {\bf B}\}$ remain fixed, we can eliminate this redundancy by caching these feature representations. Specifically, for a set of candidate inputs $\{{\bf x}_j\}_{j=1}^N$, we pre-compute and store the feature vectors:

\begin{equation}
    \bbPhi_{\text{cached}} = \{\bbphi_{{\bf W}_0+{\bf A}^\top {\bf B}}({\bf x}_n)\}_{n=1}^N
\end{equation}

When performing the recursive update in Eq.~\eqref{eq:posterior_update_1}, we directly retrieve the corresponding cached feature vector without recomputing the expensive forward pass through the LLM. This yields a modified update procedure with identical mathematical properties but significantly reduced computational complexity. The posterior mean and covariance are updated as

\begin{subequations} \label{eq:posterior_update_cached}
\begin{align}
    \hat{\bbmu}_{t+1} &= \hat{\bbmu}_{t} + \sigma_{t+1|t}^{-2}\hat{\bbSig}_{t} \bbphi_{\text{cached}}({\bf x}_{t+1})(y_{t+1} - \hat{y}_{t+1|t}) \\
    \hat{\bbSig}_{t+1} &= \hat{\bbSig}_{t} - \sigma_{t+1|t}^{-2}\hat{\bbSig}_{t} \bbphi_{\text{cached}}({\bf x}_{t+1}) \bbphi_{\text{cached}}^{\top}({\bf x}_{t+1})\hat{\bbSig}_{t}
\end{align}
\end{subequations}

where $\bbphi_{\text{cached}}({\bf x}_{t+1})$ denotes the cached feature vector for input ${\bf x}_{t+1}$ and $\hat{y}_{t+1|t} = \bbphi_{\text{cached}}^{\top}({\bf x}_{t+1})\hat{\bbmu}_t$.

This caching mechanism is particularly effective in discrete optimization domains such as molecular optimization~\cite{kristiadi2024sober}, where the search space is finite and pre-defined. The memory requirements scale linearly with the number of candidate points, making this approach feasible even for moderately large candidate sets. Furthermore, the cached features can be reused across multiple Thompson sampling iterations (cf. Eq.~\eqref{eq:AF}), amortizing the initial computation cost over the entire optimization procedure. 

The caching strategy confers a significant computational advantage over other Bayesian fine-tuning approaches that require repeated forward passes, such as Laplace approximation~\cite{yang2023bayesian} and BLoB~\cite{wang2024blob}, which do not admit such optimizations due to their non-parametric nature or joint parameter updates.

\clearpage
\bibliographystyle{IEEEtran}
\bibliography{references}

\clearpage

\end{document}